\documentclass[sigconf]{acmart}
\AtBeginDocument{%
  }

\copyrightyear{2025}
\acmYear{2025}
\setcopyright{acmlicensed}
\acmConference[CIKM '25] {Proceedings of the 34th ACM International Conference on Information and Knowledge Management}{ November 10--14, 2025}{Seoul, Republic of Korea.}
\acmBooktitle{Proceedings of the 34th ACM International Conference on Information and Knowledge Management (CIKM '25), November 10--14, 2025, Seoul, Republic of Korea}
\acmISBN{979-8-4007-2040-6/2025/11}
\acmDOI{10.1145/3746252.3761161}

\usepackage{siunitx}
\usepackage{makecell} 
\usepackage{amsmath}

\usepackage{multirow} 
\usepackage{multicol}
\usepackage{graphicx}
\usepackage{float}
\usepackage{algorithm}
\usepackage{algpseudocode}
\usepackage{balance}

\begin{document}

\title{Data-centric Prompt Tuning for Dynamic Graphs}

\author{Yufei Peng}
\authornotemark[1]
\affiliation{%
  \institution{Beijing University of Posts and Telecommunications
}
  \city{Beijing}
  \country{China}
}
\email{astral_requiem@bupt.edu.cn}

\author{Cheng Yang}
\authornote{Co-first author with equal contribution.}
\affiliation{%
  \institution{Beijing University of Posts and Telecommunications
}
  \city{Beijing}
  \country{China}
}
\email{yangcheng@bupt.edu.cn}

\author{Zhengjie Fan}
\authornotemark[2]
\affiliation{%
  \institution{Tsinghua University}
  \city{Beijing}
  \country{China}
}
\email{zjfanster@gmail.com}

\author{Chuan Shi}
\authornote{Corresponding author}
\affiliation{%
  \institution{Beijing University of Posts and Telecommunications
}
  \city{Beijing}
  \country{China}
}
\email{shichuan@bupt.edu.cn}

\renewcommand{\shortauthors}{Yufei Peng, Cheng Yang, Zhengjie Fan and Chuan Shi}

\begin{abstract}
 Dynamic graphs have attracted increasing attention due to their ability to model complex and evolving relationships in real-world scenarios. Traditional approaches typically pre-train models using dynamic link prediction and directly apply the resulting node temporal embeddings to specific downstream tasks. However, the significant differences among downstream tasks often lead to performance degradation, especially under few-shot settings. Prompt tuning has emerged as an effective solution to this problem. Existing prompting methods are often strongly coupled with specific model architectures or pretraining tasks, which makes it difficult to adapt to recent or future model designs. Moreover, their exclusive focus on modifying node or temporal features while neglecting spatial structural information leads to limited expressiveness and degraded performance. To address these limitations, we propose DDGPrompt, a data-centric prompting framework designed to effectively refine pre-trained node embeddings at the input data level, enabling better adaptability to diverse downstream tasks. We first define a unified node expression feature matrix that aggregates all relevant temporal and structural information of each node, ensuring compatibility with a wide range of dynamic graph models. Then, we introduce three prompt matrices (temporal bias, edge weight, and feature mask) to adjust the feature matrix completely, achieving task-specific adaptation of node embeddings. We evaluate DDGPrompt under a strict few-shot setting on four public dynamic graph datasets. Experimental results demonstrate that our method significantly outperforms traditional methods and prompting approaches in scenarios with limited labels and cold-start conditions.
\end{abstract}



\maketitle

\section{Introduction}

\begin{figure}[htbp]
    \centering
\includegraphics[width=0.47\textwidth]{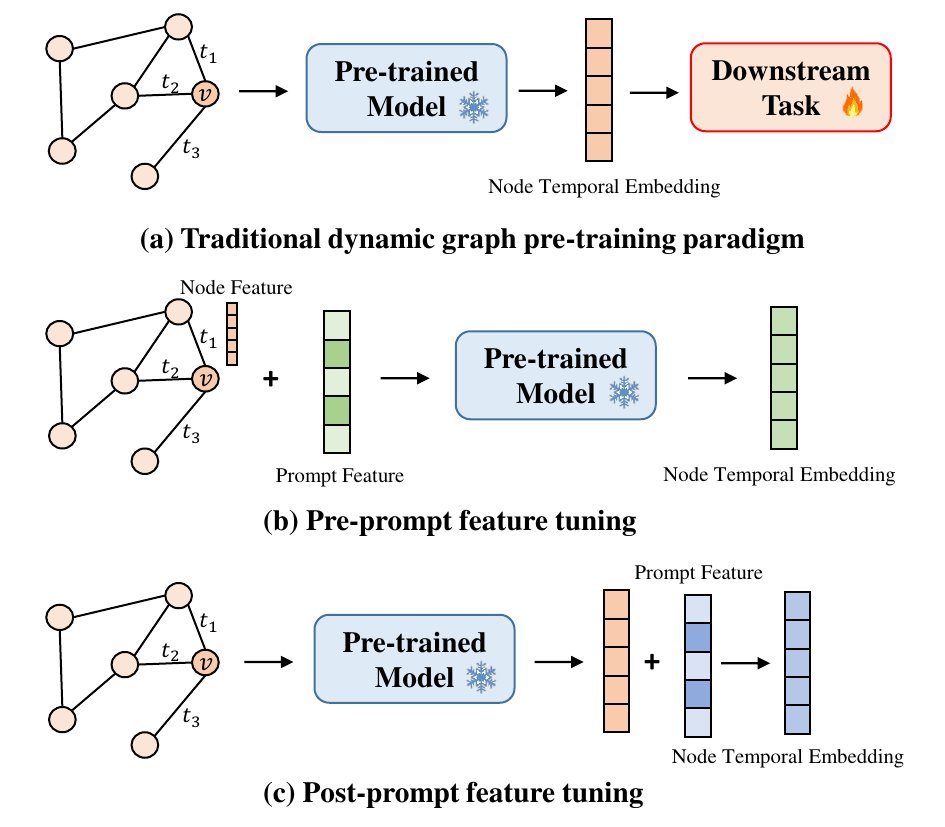}
    \caption{Comparison of (a) traditional dynamic graph methods and (b) (c) existing dynamic graph prompt works}
    \label{fig:intro}
\end{figure}

Dynamic graphs have attracted increasing attention \cite{Temporalnetworks} for their ability to model real-world systems and capture, predict, and explain time-evolving behaviors more effectively. Compared with static graphs, dynamic graphs can more effectively represent complex relationships and interactions between entities in domains such as social networks \cite{socialnetwork} and recommendation systems \cite{recommendation}, which are critical for understanding and predicting entity behavior.

Trodictional dynamic graph methods \cite{jodie,Dyrep,tgn,TGAT,graphmixer,tcl,dygformer} learn node temporal embeddings through dynamic link prediction losses using node interaction data during pretraining, and then directly apply these embeddings to downstream tasks. These methods have shown excellent performance, particularly on dynamic link prediction. Despite this progress, current dynamic graph methods still have some shortcomings. First, when adapting to various downstream tasks, traditional methods typically take one of two approaches: (1) freezing the parameters of the pre-trained model and training a lightweight decoder to map node temporal embeddings to the task; or (2) fully retraining the model for each new task. The former may introduce noise due to discrepancies between tasks, while the latter incurs considerable computational and memory costs. Second, although current methods perform well in data-rich scenarios, they often struggle in real-world applications where labeled data is scarce or where cold-start issues arise. In such few-shot or sparsely labeled settings, models that rely heavily on large-scale labeled data typically fail to generalize. 

In order to solve the above problems in dynamic graphs, some recent approaches have explored prompt tuning~\cite{tigprompt,DyGPrompt}, which has proven effective in the field of static graphs~\cite{gppt,Graphprompt,gpf,All-in-one}. As shown in Fig.~\ref{fig:intro}, these methods introduce lightweight prompt modules that adjust graph input features or output embeddings with minimal additional parameters. Specifically, TIGPrompt~\cite{tigprompt} incorporates recent interaction features as post-prompt added to pre-trained node temporal embeddings. DyGPrompt~\cite{DyGPrompt} applies a unified learnable embedding to all node and time features, further fine-tuned by pre-prompt based on the interaction of node and time features.  

However, recent prompting methods for dynamic graphs still suffer from two critical limitations. First, existing approaches often require intervention into model-specific structures or adopt entirely different pretraining and inference paradigms from prior works. For instance, DyGPrompt aligns downstream tasks with link-based pretraining objectives through its prompting mechanism, deviating from the common paradigm where node embeddings are directly fed into task-specific classifiers. This design makes it difficult to integrate DyGPrompt into existing models and may result in significant performance loss. Additionally, both TIGPrompt and DyGPrompt focus solely on modifying node or time features, completely ignoring the spatial neighborhood structure that is intrinsic to dynamic graphs. This under-expressive prompting strategy hampers their ability to comprehensively capture task-relevant patterns, ultimately constraining downstream performance.

To address these two challenges, we propose DDGPrompt, a novel data-centric prompting framework for dynamic graphs. DDGPrompt adjusts the input data structure in a unified and model-agnostic manner, enabling better alignment between pre-trained models and downstream tasks. Specifically, to tackle the first challenge, we define a node expression feature matrix inspired by existing dynamic graph models. This matrix is compatible with a wide range of traditional architectures and designed to remain extensible to future frameworks. It encodes a node’s recent first-order neighborhood information, including node features, edge features, and time features, and serves as input to various backbone models to get node temporal embeddings. To address the second challenge, we introduce three complementary prompt matrices (temporal bias, edge weight, and feature mask) to comprehensively adjust the node expression feature matrix. The temporal bias prompt dynamically adjusts the temporal features of each neighbor; the edge weight prompt assigns a learnable importance score to each neighbor, capturing spatial structural relevance; and the feature mask prompt leverages a task-aware enhancement network to selectively modulate feature dimensions. These prompts are integrated into the node expression matrix through different weights, allowing the resulting temporal embeddings to be more expressive and better adapted to diverse downstream tasks.

Finally, we conduct extensive experiments on four public dynamic graph datasets. Thanks to our proposed node expression feature matrix and prompting strategy, DDGPrompt achieves strong performance across various challenging few-shot scenarios. It is compatible with existing dynamic graph models and consistently outperforms both backbone baselines and recent prompt-based methods.

Our contributions are summarized as follows:

\(\bullet\) We define a node expression feature matrix for efficient prompt tuning in dynamic graphs. This matrix encodes the historical interaction information of each node and maintains compatibility with all existing methods. It serves as the model input to generate node temporal embeddings that are transferable across different downstream tasks.

\(\bullet\) We propose a novel data-centric prompting framework, DDGPrompt, which integrates temporal bias, edge weight, and feature mask prompts. By jointly capturing node, temporal, and spatial information, it adaptively refines the node expression matrix for task-specific tuning.

\(\bullet\) Extensive few-shot experiments on four dynamic graph datasets demonstrate that DDGPrompt consistently outperforms various backbones and prompt methods, highlighting its robustness and effectiveness.

\section{Related work}

\subsection{Dynamic Graph Learning}

Dynamic graphs can be categorized into discrete-time dynamic graph (DTDG) and continuous-time dynamic graph (CTDG) based on the granularity of their temporal information \cite{dynamicgraph_survey,kazemi2020representation,barros2021survey}. In recent years, CTDGs have attracted increasing attention due to their closer alignment with real-world scenarios \cite{jodie,tgn,TGAT,graphmixer,tcl,dygformer}. Most existing CTDG methods share a common architecture: they design a backbone encoder to extract spatio-temporal information from the evolving graph and obtain node representations, followed by a decoder that leverages these embeddings to perform downstream tasks. For instance, TGN \cite{tgn} introduces a memory module to store and update node embeddings over time. GraphMixer \cite{graphmixer} employs a lightweight MLP-based architecture that significantly reduces memory and computation cost while maintaining competitive performance. TCL \cite{tcl}, built upon contrastive learning, uses a transformer-based backbone to capture temporal dynamics. Our work also focuses on the CTDG setting.

While these methods have achieved perfect performance on dynamic link prediction, they often fall short when applied to other downstream tasks. This limitation is particularly pronounced in dynamic graphs due to inherent temporal gaps and node preference variability over time, leading to even greater performance degradation across tasks compared to static graphs. Furthermore, most existing approaches overlook few-shot and cold-start scenarios. In practical settings where labeled data is scarce or user-item interactions are sparse, such models tend to generalize poorly and lack robustness.

\subsection{Graph pretraining}

Pretraining has emerged as a prevalent paradigm in graph learning \cite{pre-training,Gcc,lu2021learning,jiang2021pre, hu2020gpt}. By training models on large-scale datasets and then fine-tuning them for specific downstream tasks, pretraining significantly boosts performance. This paradigm leverages the power of transfer learning \cite{transfer,zhuang2020comprehensive,niu2020decade}, allowing models to generalize better by reusing the knowledge acquired during pretraining.

Recent dynamic graph learning methods have begun to adopt pretraining strategies to enable transferability across multiple tasks \cite{dygformer}. Typically, these methods perform supervised pretraining using dynamic link prediction task. When adapting to new downstream tasks, they freeze the parameters of the pre-trained model and retrain only a task-specific classifier based on the node temporal embeddings produced during pretraining.

However, this approach has inherent limitations. Since the temporal embeddings are optimized for a specific task during supervised pretraining, they may encode task-biased information, resulting in suboptimal generalization to other downstream tasks. To address this issue, it is critical to incorporate task-agnostic self-supervised signals during pretraining. It can enhance the versatility of node embeddings, improve performance across diverse downstream tasks, and offer greater robustness in low-resource or few-shot scenarios where labeled data are scarce.

\subsection{Prompt Learning on Graphs}
In the past two years, prompt learning has been increasingly applied to the graph learning domain and has achieved remarkable success \cite{gppt,All-in-one,Graphprompt,gpf,MultiGPrompt,aagod,Hgprompt,sun2023graph}. These studies explore ways to align diverse downstream tasks with pre-trained graph models and have shown strong performance, particularly in few-shot settings. Some works even aim to unify various tasks and datasets under a single graph foundation model \cite{MultiGPrompt,Opengraph,liu2023one,he2024unigraph,foundation_models}. Existing prompt learning approaches for static graphs can be broadly categorized into two types \cite{Data-centric}: (1) \textit{Pre-prompt} usually modifies the model input. This idea is analogous to the prompt in natural language processing. For instance, GPF \cite{gpf} proposes universal feature prompts for all nodes, while All-in-one \cite{All-in-one} explicitly alters the graph structure to encode task-specific signals. (2) \textit{Post-prompt} usually uses the prompt vector to modify the node embedding obtained by the model to adapt to different downstream tasks. For example, GraphPrompt \cite{Graphprompt} introduces a prompt tensor that multiplies the node embeddings, enabling the generation of task-specific node representations via a unified prompt template. Beyond general graph tasks, prompt learning has also been applied to more specialized settings \cite{aagod,dcgc}.

Recently, a few works have extended prompt learning to dynamic graphs \cite{tigprompt,DyGPrompt}. TIGPrompt \cite{tigprompt} injects recent interaction history as prompt signals to enhance node embeddings. DyGPrompt \cite{DyGPrompt}, inspired by GraphPrompt, designs global feature prompts for both node and time features while incorporating their interactions. However, despite these efforts, such methods often struggle to generalize across different datasets due to the limited expressiveness of the learned prompt representations.

\section{Preliminary}

\textbf{Definition 1. Dynamic Graph} Dynamic graph can be represented as \(G = (V,E)\) \cite{defination}. \(V\) and \(E\) are denoted as the node set and the edge set, respectively. Each edge consists of a triple \(e = (u,v,t) \in E\), where \(u \in V\) is the source node, \(v \in V\) is the destination node, and \(t\) is the timestamp, indicating that \(u\) and \(v\) interact at timestamp \(t\).

The features corresponding to nodes \(u,v\) and edge \(e\) are denoted as \(f_{u}, f_{v} \in \mathbb{R}^{d_N}\) and \(f_{e} \in \mathbb{R}^{d_E}\), respectively. \(d_N\) and \(d_E\) are feature dimensions.

\textbf{Definition 2. Problem Formalization.}

\textit{Dynamic link prediction:} For a pair of nodes \(u\) and \(v\) in a dynamic graph \(G\) and a given time \(t\), the purpose of dynamic link prediction is to predict whether there will be a link between the two nodes at time \(t\) based on all interactions \(\{(u', v', t') | t' < t\}\) before the timestamp.

\textit{Dynamic node classification:} For each edge \(e = (u, v, t)\) in the dynamic graph \(G\), source node \(u\) corresponds to a label \(l \in L \) at timestamp \(t\), where the labels of all source nodes constitute the label set \(L\). The goal of dynamic node classification is to predict the label of the source node \(u\) at timestamp \(t\).

In this paper, we focus on two fundamental tasks on dynamic graphs under the few-shot setting, as real-world dynamic graphs often suffer from limited supervision signals \cite{Melu,Warm,Meta_gnn,yao2020graph}.

\section{Methodology}
\subsection{Overall Framework}

\begin{figure*}[htbp]
    \centering
\includegraphics[width=0.9\textwidth]{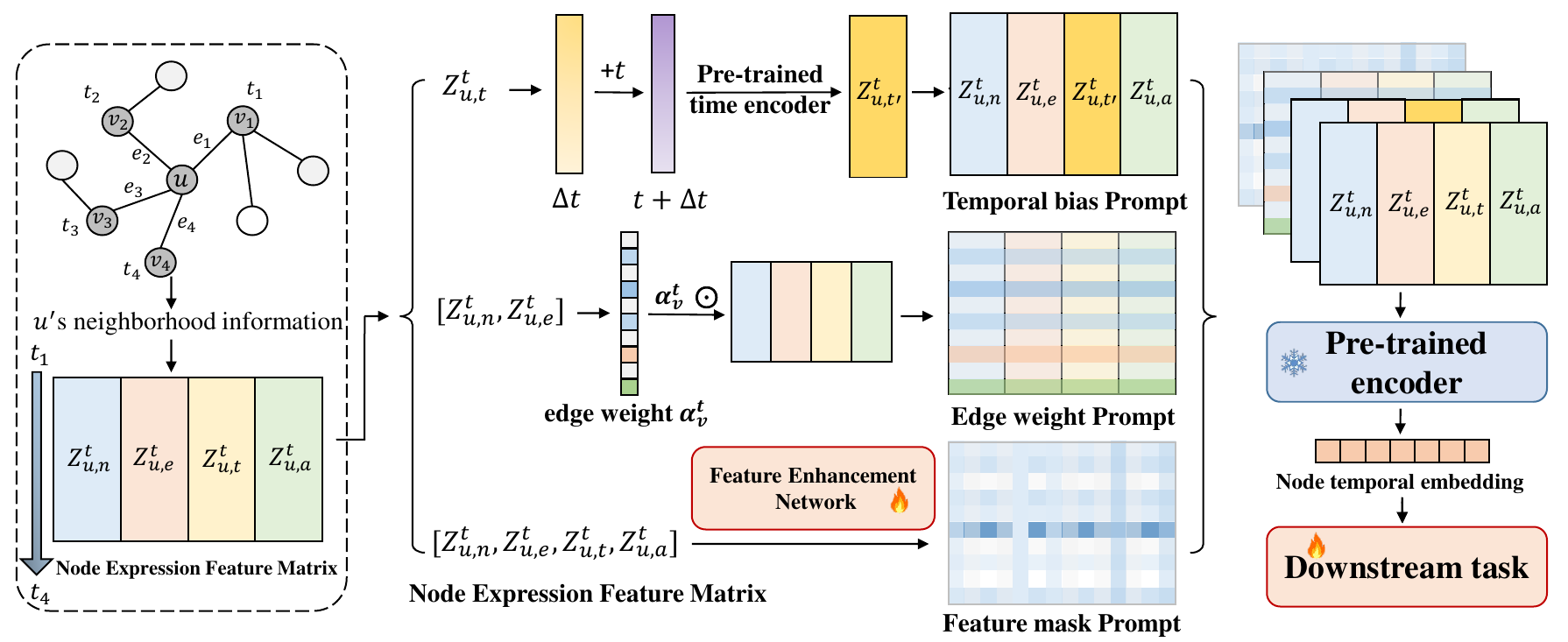}
    \caption{The overall framework of our proposed DDGPrompt}
    \label{fig:method}
\end{figure*}

The overall framework consists of two parts: pretraining and downstream task fine-tuning. In the pretraining stage, we first train the model using a large amount of unlabeled data through self-supervised contrastive learning~\cite{oord2018representation} based on link prediction. Afterwards, we first define a node expression feature matrix for each node. This matrix is then adjusted using three prompt matrices. The result is used as the input of the pre-trained model with frozen parameters, and the node temporal embedding modified by prompt is obtained for different downstream tasks. The overall framework of DDGPrompt is shown in Fig. \ref{fig:method}.

\subsection{Node Expression Feature Matrix}
\label{sec:Feature_Matrix}

In this section, we first define a node expression feature matrix for each node based on the common design patterns of existing dynamic graph models, aiming to optimize node temporal embeddings through prompt tuning. Previous works \cite{graphmixer,tcl,dygformer} have shown that it is sufficient to capture the general characteristics of a node through its first-order interactions in its recent history. Specifically, for an interaction \((u, v, t)\), we take node \(u\) as an example and extract the most recent sequence of interactions \(S_u^t = \{(u, v', t') | t' < t\}\in G\) of node \(u\) from the interaction history. The original features of all neighbor nodes are \(\textbf{F}^t_{u,neigh} \in \mathbb{R}^{|S_u^t|\times d_N} \). Consequently, the edge features and the time features are \(\textbf{F}^t_{u,edge} \in \mathbb{R}^{|S_u^t|\times d_E} \) and \(\textbf{F}^t_{u,time} \in \mathbb{R}^{|S_u^t|\times d_T} \). For time features, previous works \cite{TGAT,graphmixer} use time encoding function \(f_{time}(\Delta t ; \omega)\) to encode the time interval \(\Delta t = t - t'\) into T-dimensional time features \(\textbf{F}^t_{u,time}\), where \(\omega\) is the parameter of the time encoding function.

Next, we use a linear layer to project each original feature into the same feature space so that they have the same dimension \(d\) and obtain the embedding of each feature.

\begin{gather}
    \textbf{Z}^t_{u,neigh} = \text{Projection}(\textbf{F}^t_{u,neigh})\in 
    \mathbb{R}^{|S_u^t|\times d}\\
    \textbf{Z}^t_{u,edge} = \text{Projection}(\textbf{F}^t_{u,edge})\in 
    \mathbb{R}^{|S_u^t|\times d}\\
    \textbf{Z}^t_{u,time} = \text{Projection}(\textbf{F}^t_{u,time})\in 
    \mathbb{R}^{|S_u^t|\times d}
    \label{eq:projection}
\end{gather}

We define the node expression feature matrix of node \(u\) at timestamp \(t\) as follows:

\begin{equation}
    \textbf{Z}^t_u = \textbf{Z}^t_{u,neigh} || \textbf{Z}^t_{u,edge} || \textbf{Z}^t_{u,time} \in 
    \mathbb{R}^{|S_u^t|\times 3d}
\end{equation}

where \(||\) represents the concatenate operation. 

In addition, for some specific methods, they may propose some additional features to improve the performance of specific tasks. For example, DyGFormer \cite{dygformer} proposes a co-occurrence feature \(\textbf{Z}^t_{u,occ} \in \mathbb{R}^{|S_u^t|\times d}\) between a pair of nodes to extract the implicit information. We will collectively denote this additional features as \(\textbf{Z}^t_{u,add}\). If \(\textbf{Z}^t_{u,add}\) exists, we add it to the end as \(\textbf{Z}^t_u || \textbf{Z}^t_{u,add}\). This makes it compatible with existing methods and directly extensible to future work. Obviously, the matrix \(\textbf{Z}^t_u\) is a combination of features for each historical neighbor of the node \(u\) at timestamp \(t\).

Afterwards, the feature matrix \(\textbf{Z}^t_u\) will be used as input to different backbone models to extract the information from node \(u\) and output the node temporal embedding \(h^t_u\).

\subsection{Self-supervised pretraining}

We leverage the node expression feature matrix introduced in Section~\ref{sec:Feature_Matrix} to pretrain the backbone model using a large amount of unlabeled dynamic graph data, resulting in the initial node temporal embedding\(h^t_u\). To optimize the pre-trained model, we adopt a self-supervised contrastive learning loss based on link prediction, as the dynamic interactions between nodes inherently capture rich temporal and structural information from the graph.

Specifically, for an interaction \((u, v, t) \in G_{pre-train}\) of node \(u\) in the dynamic graph \(G\), where \(G_{pre-train}\) is the pretraining dataset split on \(G\). We randomly sample another node \(v-\), which forms a triple \((u, v-, t)\) with \(u\) as a negative sample, indicating that there is no edge between \(u\) and \(v\) up to the timestamp \(t\). We use contrastive learning to make the embeddings of two nodes with a link relationship similar and the embeddings of two nodes without a link relationship far away from each other. Therefore, for each interaction, we use the following loss function to optimize the pre-trained model:
\begin{equation}
    \mathcal{L}_{\text{pre}}(\Phi) = -\sum_{(u, v, t) \in G_{pre-train}} \ln \frac{e^{\frac{1}{\tau} \text{sim}(h^t_{u}, h^t_{v})}}{{e^{\frac{1}{\tau} \text{sim}(h^t_{u}, h^t_{v})}} + e^{\frac{1}{\tau} \text{sim}(h^t_{u}, h^t_{v-})}}
    \label{eq:pre_loss}
\end{equation}

where \(\Phi\) is the trainable parameter of the pre-trained backbone. \(\tau\) is the temperature parameter. \({h}^t_{u},{h}^t_{v},{h}^t_{v-}\) is the temporal embedding of the node at timestamp \(t\) for downstream tasks, which is calculated by the pre-trained model with \(\Phi\). \(\text{sim}(\cdot)\) is used to measure the similarity between node embeddings, and we use the cosine similarity.

\subsection{Data-centric Dynamic Graph Prompt}

To bridge the gap between pre-trained embeddings and downstream tasks, we introduce three types of prompts that refine the node expression feature matrix from node, temporal, and spatial perspectives. These prompts adjust the resulting temporal embeddings, enabling the pre-trained model to adapt to downstream tasks more efficiently and effectively.

\subsubsection{Temporal bias prompt matrix.} 
The relationship between the node attribute and its time feature is a key point that distinguishes dynamic graphs from traditional graphs. However, prior methods often model the features of neighboring nodes and their corresponding timestamps independently, overlooking the crucial interactions between them~\cite{TGAT,graphmixer,dygformer}.  In fact, enhancing the time features based on the neighbor node feature can more accurately capture the behavioral preferences of the current node. For instance, adjusting a neighbor’s interaction timestamp to be closer to the current time may indicate a stronger preference or relevance to that neighbor. To capture this intuition, we propose a temporal bias prompt that explicitly modifies the time feature to reflect such dynamics.

We first extract the neighbor node feature \(\mathbf{Z}^t_{u,neigh}\) of the node expression feature matrix and generate a one-dimensional temporal bias \(\delta t\) through a linear layer as the time gap that needs to be modified for different neighbor nodes.

\begin{equation}
    \delta t = \text{Linear}(\mathbf{Z}^t_{u,time} ; \eta)
\end{equation}

 where \(\eta\) is the learnable parameter of the linear layer. Subsequently, we modified the the original time interval \(\Delta t\) using temporal bias \(\delta t\) so that the interaction time of neighbor are adjusted to either a more recent or more distant point in time. We use ReLU as the activation function to prevent the time interval from being less than 0.

\begin{equation}
    {{\overline{\Delta t}}} = \text{ReLu}(\Delta t + \delta {t})
\end{equation}

The modified time interval \({{\overline{\Delta t}}}\) will be used as input to obtain new time features \(\overline{\mathbf{Z}}^t_{u,time}\) using the time encoding function and projection layer of the pre-trained model as before.

\begin{equation}
    \overline{\mathbf{Z}}^t_{u,time} = \text{Projection}(\text{TimeEncoder}({\overline{\Delta t}}))
\end{equation}

We use \(\overline{\mathbf{Z}}^t_{u,time}\) to replace the original time feature and obtain the temporal bias prompt matrix \(\mathbf{P}_{temp}\)of the same size as the original node expression feature matrix.

\begin{equation}
    \mathbf{P}_{temp} = \mathbf{Z}^t_{u,neigh} || \mathbf{Z}^t_{u,edge} || \overline{\mathbf{Z}}^t_{u,time} \in 
    \mathbb{R}^{|S_u^t|\times 3d}
    \label{eq:time_prompt}
\end{equation}

\subsubsection{Edge weight prompt matrix.}
Previous traditional and prompt methods overlook the varying importance of different neighbors when aggregating neighborhood information. They typically assign equal weights to all neighbors \cite{tcl,graphmixer,dygformer,tigprompt,DyGPrompt}. However, this uniform treatment becomes a critical limitation when adapting to different downstream tasks, often resulting in significant performance degradation. For example, in dynamic link prediction, it is beneficial to assign higher weights to frequently interacting node pairs, whereas dynamic node classification may prioritize the most recent interactions. Therefore, capturing task-specific neighbor importance is essential for improving generalization across diverse tasks.

Therefore, we generate the edge weights according to the different neighbor node features and edge features to adaptively adjust the information expression on different edges. Specifically, for node \(u\)'s neighbor node \(v\) at timestamp \(t\), we concatenate the neighbor node features \(z^t_{uv,neigh}\) and edge features \(z^t_{uv,edge}\) and use them as the input of a linear layer to get the edge weight \(w^t_{uv}\).

\begin{equation}
    w^t_{uv} = \text{Linear}(z^t_{uv,neigh} || z^t_{uv,edge} ; \zeta)
\end{equation}

where \(\zeta\) is the learnable parameter of the linear layer. The weight of all edges is denoted by \(w^t_u\). We then multiply the weight by the corresponding edge.

\begin{equation}
    \mathbf{P}_{edge} = w^t_u \odot \mathbf{Z}^t_u
    \label{eq:edge_prompt}
\end{equation}

where \((\odot)\) is the element-wise multiplication. It modifies all feature information on an edge through a broadcast mechanism. Finally, we get the edge weight prompt matrix \(\mathbf{P}_{edge}\).

\subsubsection{Feature mask prompt matrix.}
Finally, we introduce a feature mask prompt matrix to enhance the representation of each neighbor's features. The goal is to adaptively refine the original node feature matrix through a Feature Enhancement Network (FEN). To keep the architecture lightweight, we employ a two-layer MLP with a bottleneck structure. The enhancement process is defined as follows:

\begin{equation}
    \textbf{P}_{feat} = \text{MLP}(\textbf{Z}^{t}_u ; \Omega) = \mathbf{W_2}\cdot\text{Relu}(\mathbf{W_1}\cdot\mathbf{Z}^{t}_u + b_1) + b_2
    \label{eq:feature_prompt}
\end{equation}

Here, \(\mathbf{W}_1, \mathbf{W}_2\) and \(b_1, b_2\) denoted as \(\Omega\) are the learnable parameters of the MLP. The output \(\textbf{P}_{feat} \) also maintains the same dimensionality as the original feature matrix, ensuring compatibility for subsequent operations. This design is also commonly adopted in traditional static graph learning. It does not assume any specific structure or distribution of node features, making it applicable to a wide range of graph-based tasks.

Finally, we add the three prompt matrices to the original node expression feature matrix with different weights.

\begin{equation}
    \mathbf{\overline{Z}}^t_{u} =
     \mathbf{Z}^t_{u} + \alpha \odot\mathbf{P}_{temp} + \beta \odot \mathbf{P}_{edge} +\gamma \odot \mathbf{P}_{feat}
    \label{eq:add_all_prompt}
\end{equation}

\(\alpha,\beta,\gamma\) are the weight hyperparameters corresponding to the three prompts, which are used to further adjust the expression of the three prompts on the original feature matrix. 
\(\mathbf{\overline{Z}}^t_{u}\) is the prompt-adjusted node expression feature matrix. It is treated as input and passes through pre-trained backbone \(\text{Pre-train}(\mathbf{\Phi})\) to generate new node temporal embedding \(\overline{h}^{t}_u\).

\begin{equation}       
    {\overline{h}}^{t}_u = \text{Pre-train}(\mathbf{\overline{Z}}^t_{u} ; \Phi)
\end{equation}


\subsection{Downstream Task Tuning}

In line with previous methods \cite{TGAT,tcl,dygformer}, we finally train MLP as a classifier to apply prompt-adjusted node temporal embeddings to different downstream tasks. Both dynamic link prediction and dynamic node classification are formulated as binary classification problems, and we adopt the binary cross-entropy (BCE) loss function for optimization. Additionally, a regularization term is introduced for the FEN to mitigate overfitting. For dynamic link prediction, we apply the following loss on the downstream training set \(G_{fine-tune} \in G\), optimizing only the prompt parameters and the classifier, while keeping the backbone frozen. Here, \(l\) denotes the ground-truth label in \(G_{fine-tune}\).

\begin{equation}
    \mathcal{L}_{\text{link}}(\eta,\zeta,\Omega,\Theta) = \text{Cross-Entropy}(\text{MLP}(\overline{h}^{t}_u||\overline{h}^{t}_v;\Theta),l) + \lambda ||\Omega||^2
    \label{eq:link_loss}
\end{equation}

\(\eta,\zeta,\Omega\) are the trainable parameters for prompt. \(\Theta\) is the trainable parameters of the MLP. \(\lambda\) represents the strength of regularization, and \(||\cdot||\) is L2-norm.


Similarly, the loss for dynamic node classification is as follows:

\begin{equation}
    \mathcal{L}_{\text{node}}(\eta,\zeta,\Omega,\Theta) = \text{Cross-Entropy}(\text{MLP}(\overline{h}^{t}_u;\Theta),l) + \lambda ||\Omega||^2
    \label{eq:node_loss}
\end{equation}

\begin{algorithm}
\caption{DDGPrompt Pretraining and Fine-tuning Framework}
\label{alg:recommendation}
\begin{algorithmic}[1]

\Require Dynamic graph pretraining dataset \(G_{pre-train}\) and fine-tuning dataset \(G_{fine-tune}\), node features set \(F_{N}\), edge features set \(F_{E}\), label set label \(L\), backbone model \(f_{\Phi}\) with parameters \(\Phi\)

\State // Pretraining Stage:
\For{each edge \(e = (u, v, t) \in G_{pre-train} \)}
    \State for node \(u\), extract the features \(\textbf{F}^t_{u,neigh},\textbf{F}^t_{u,edge},\textbf{F}^t_{u,time}\) of the node's one-hop neighbors
    \State Calculate project features \(\textbf{Z}^t_{u,neigh} ,\textbf{Z}^t_{u,edge},\textbf{Z}^t_{u,time}\) via Eq.~\ref{eq:projection}
    \State \(\textbf{Z}^t_u \leftarrow Contact( \textbf{Z}^t_{u,neigh}, \textbf{Z}^t_{u,edge},\textbf{Z}^t_{u,time} )\)
    \State Sample negetive node and get \(\textbf{Z}^t_v,\textbf{Z}^t_{v-}\) similarity
    \State Calculate embeddings \({h}^t_{u},{h}^t_{v},{h}^t_{v-}\) from \(f_{\Phi}\)
    \State Calculate loss \(\mathcal{L}_{\text{pre}}(\Phi)\) and optimize model paremeter \(\Phi\)
\EndFor

\State // Fine-tuning Stage:

\For{each edge \(e = (u, v, t) \in G_{fine-tune} \)}
  \State \(\delta t \leftarrow \text{Linear}(\mathbf{Z}^t_{u,time} ; \eta),{{\overline{\Delta t}}} \leftarrow \text{ReLu}(\Delta t + \delta {t})\)
  \State Calculate time bias prompt \(\mathbf{P}_{temp}\) via Eq.~\ref{eq:time_prompt}
  \State \(w^t_u \leftarrow Linear(\textbf{Z}^t_{u,neigh} ||\textbf{Z}^t_{u,edge} ; \zeta)\)
  \State Calculate edge weight prompt \(\mathbf{P}_{edge}\) via Eq.~\ref{eq:edge_prompt}
  \State Calculate feature mask prompt \(\textbf{P}_{feat} \leftarrow \text{MLP}(\textbf{Z}^{t}_u ; \Omega)\) via Eq.~\ref{eq:feature_prompt}
  \State \(\overline{Z}^t_{u} \leftarrow Fine-tuning(\mathbf{P}_{neigh},\mathbf{P}_{edge},\mathbf{P}_{feat}) \) via Eq.~\ref{eq:add_all_prompt}
  \State Calculate embeddings \(\overline{h}^t_{u},\overline{h}^t_{v},\overline{h}^t_{v-}\) from freezed pre-trained model \(f_{\Phi}\)
  \State Calculate loss \(\mathcal{L}_{\text{link}}\) or \(\mathcal{L}_{\text{node}}\)  and optimize prompt paremeter \(\eta,\zeta,\Omega\) and MLP classifier \(\Theta\) via Eq.~\ref{eq:link_loss} or Eq.~\ref{eq:node_loss}
  
\EndFor
\end{algorithmic}
\end{algorithm}

\section{Experiment}

In this section, we conduct extensive experiments to evaluate the performance of our model in the scenario where labeled data is scarce.

\subsection{Experimental Setup}

\subsubsection{Datasets}
We evaluate our approach and baselines using four widely used benchmark datasets (Wikipedia, Reddit, MOOC, and a large-scale dataset LastFM) on dynamic graphs. The detailed description of the dataset is as follows:

\begin{table}[h]
    \centering
    \caption{Detailed
statistics of Datasets}
    \label{tab:datasets}
    \begin{tabular}{lcccc}
        \toprule
        Datasets & Wikipedia & Reddit & MOOC & LastFM \\ \midrule
        Nodes & 9227 & 11000 & 7144 & 1980 \\
        Edges & 157474 & 672447 & 411749 & 1293103 \\
        Feature dimension & 172 & 172 & 172 & 0 \\
        Node classes nums & 2 & 2 & 2 & 0 \\
        Dynamic labels & 217 & 366 & 4066 & 0 \\
        Timespan & 30 days & 30 days & 30 days & 30 days \\ \bottomrule
    \end{tabular}
\end{table}

\begin{table*}[h]
    \renewcommand{\arraystretch}{1.1}
    \centering
    \caption{Experimental results of few-shot scenario}
    \label{tab:main_shiyan}
    \resizebox{\textwidth}{!}{
    \begin{tabular}{l|cccccccc|cccccccc|ccc}
        \toprule
        \multirow{2}{*}{\textbf{Methods}} &
        \multicolumn{8}{c|}{\textbf{Transductive Link Prediction}} & 
        \multicolumn{8}{c|}{\textbf{Inductive Link Prediction}} & 
        \multicolumn{3}{c}{\textbf{Node Classification }} \\
        & 
        \multicolumn{2}{c}
        {\textbf{Wikipedia}} & 
        \multicolumn{2}{c}
        {\textbf{Reddit}} & 
        \multicolumn{2}{c}
        {\textbf{MOOC}} & 
        \multicolumn{2}{c|}
        {\textbf{LastFM}} &
        \multicolumn{2}{c}
        {\textbf{Wikipedia}} & 
        \multicolumn{2}{c}
        {\textbf{Reddit}} & 
        \multicolumn{2}{c}
        {\textbf{MOOC}} & 
        \multicolumn{2}{c|}
        {\textbf{LastFM}} &
        \textbf{Wikipedia} & \textbf{Reddit} &
        \textbf{MOOC} \\
        
        \midrule
        \textbf{Metrics} &
        \textbf{AP$\uparrow$} &
        \textbf{AUC$\uparrow$} &
        \textbf{AP$\uparrow$} &
        \textbf{AUC$\uparrow$} &
        \textbf{AP$\uparrow$} &
        \textbf{AUC$\uparrow$} &
        \textbf{AP$\uparrow$} &
        \textbf{AUC$\uparrow$} &
        \textbf{AP$\uparrow$} &
        \textbf{AUC$\uparrow$} &
        \textbf{AP$\uparrow$} &
        \textbf{AUC$\uparrow$} &
        \textbf{AP$\uparrow$} &
        \textbf{AUC$\uparrow$} &
        \textbf{AP$\uparrow$} &
        \textbf{AUC$\uparrow$} &
        \textbf{AUC$\uparrow$} &
        \textbf{AUC$\uparrow$} &  \textbf{AUC$\uparrow$} \\

        \midrule
        JODIE & 
        52.26 &
        50.22 &
        56.28 &
        55.08 &
        49.15 &
        49.29 &
        49.09 &
        45.88 &
        52.86 &
        50.52 &
        58.54 &
        55.35 &
        47.66 &
        47.51 &
        48.74 &
        42.84 &
        \underline{72.93} &
        57.12 &
        52.42 \\
        
        TGN & 
        58.67 &
        56.63 &
        57.45 &
        54.34 &
        53.92 &
        55.39 &
        54.17 &
        53.81 &
        58.64 &
        56.41 &
        53.36 &
        49.59 &
        54.13 &
        54.39 &
        50.73 &
        49.70 &
        41.54 &
        51.86 &
        49.20 \\
        
        TGAT & 
        73.98 &
        71.32 &
        63.83 &
        64.43 &
        69.19 &
        68.99 &
        59.20 &
        51.88 &
        75.80 &
        73.65 &
        61.31 &
        62.84 &
        67.93 &
        69.12 &
        53.82 &
        52.73 &
        61.92 &
        57.38 &
        49.17 \\
        
        GraphMixer &
        74.63 &
        73.52 &
        81.13 &
        79.88 &
        72.03 &
        71.19 &
        51.40 &
        50.33 &
        74.43 &
        73.70 &
        78.25 &
        77.78 &
        72.19 &
        72.03 &
        51.40 &
        50.27 &
        54.53 &
        55.01 &
        57.82 \\
        
        TCL & 
        80.37 &
        74.57 &
        87.54 &
        88.77 &
        \underline{78.14} &
        \underline{79.65} &
        52.81 &
        50.82 &
        80.42 &
        74.90 &
        86.14 &
        87.57 &
        \underline{77.30} &
        \underline{79.28} &
        52.72 &
        50.81 &
        61.71 &
        54.01 &
        52.56 \\

        DyGFormer & 
        85.78 &
        83.03 &
        97.66 &
        97.50 &
        70.94 &
        69.49 &
        64.83 &
        66.44 &
        86.93 &
        83.58 &
        97.97 &
        97.56 &
        68.74 &
        67.17 &
        64.94 &
        66.72 &
        63.67 &
        \underline{60.07} &
        54.45 \\

        \midrule

        TIGPrompt &
        58.35 &
        56.64 &
        54.72 &
        51.62 &
        54.76 &
        55.30 &
        53.22 &
        52.77 &
        57.73 &
        55.93 &
        51.57 &
        47.33 &
        54.91 &
        54.65 &
        50.60 &
        49.51 &
        46.13 &
        53.85 &
        56.46 \\

        DyGPrompt &
        50.65 &
        49.56 &
        55.41 &
        52.49 &
        52.92 &
        54.83 &
        48.97 &
        47.05 &
        50.86 &
        49.90 &
        52.03 &
        49.14 &
        54.23 &
        54.34 &
        48.66 &
        46.46 &
        46.16 &
        49.60 &
        48.98 \\

        \midrule
        \midrule




        
        TIGPrompt(T) &
        81.32 &
        76.08 &
        85.99 &
        86.19 &
        77.66 &
        78.83 &
        54.39 &
        52.32 &
        81.37 &
        76.36 &
        83.73 &
        85.09 &
        76.62 &
        78.15 &
        54.33 &
        52.43 &
        59.34 &
        55.16 &
        \underline{59.13} \\

        DyGPrompt(T) &
        79.89 &
        77.53 &
        83.80 &
        86.50 &
        53.30 &
        55.24 &
        51.32 &
        50.67 &
        79.15 &
        76.58 &
        82.33 &
        84.29 &
        53.17 &
        55.76 &
        51.37 &
        50.75 &
        75.03 &
        55.42 &
        42.08 \\

        \textbf{DDGPrompt(T)} &
        88.06 &
        84.23 &
        86.35 &
        87.57 &
        \textbf{79.41} &
        \textbf{80.96} &
        50.59 &
        49.63 &
        87.42 &
        83.49 &
        84.81 &
        86.32 &
        \textbf{78.47} &
        \textbf{80.43} &
        50.53 &
        49.58 &
        66.24 &
        55.99 &
        49.62 \\
        
        \midrule
        
        TIGPrompt(D) &
        84.45 &
        81.83 &
        \underline{98.45} &
        \underline{98.22} &
        69.88 &
        68.49 &
        62.11 &
        63.96 &
        85.78 &
        82.47 &
        \underline{98.54} &
        \underline{98.32} &
        67.61 &
        65.89 &
        62.14 &
        64.17 &
        65.96 &
        53.82 &
        57.20 \\

        DyGPrompt(D) &
        \underline{92.76} &
        \underline{90.33} &
        60.19 &
        58.86 &
        67.17 &
        67.08 &
        \underline{72.25} &
        \underline{69.23} &
        \underline{92.85} &
        \underline{90.38} &
        60.06 &
        58.72 &
        65.58 &
        66.02 &
        \underline{72.46} &
        \underline{68.96} &
        67.81 &
        57.35 &
        45.12 \\

        \textbf{DDGPrompt} &
        \textbf{97.15} &
        \textbf{96.32} &
        \textbf{98.53}&
        \textbf{98.36} &
        72.22 &
        71.03 &
        \textbf{80.48} &
        \textbf{76.29} &
        \textbf{96.91} &
        \textbf{96.01} &
        \textbf{98.59} &
        \textbf{98.41} &
        70.76 &
        69.67 &
        \textbf{80.53} &
        \textbf{76.28} &
        \textbf{74.32} &
        \textbf{61.30} &
        \textbf{59.60} \\
        \bottomrule
    \end{tabular}
    }
\end{table*}

\begin{table*}[h]
    \renewcommand{\arraystretch}{1.1}
    \centering
    \caption{ Experimental results of sparse interaction data scenario}
    \label{tab:few_interact_exp}
    \resizebox{\textwidth}{!}{
    \begin{tabular}{l|cccccccc|cccccccc|ccc}
        \toprule
        \multirow{2}{*}{\textbf{Methods}} &
        \multicolumn{8}{c|}{\textbf{Transductive Link Prediction}} & 
        \multicolumn{8}{c|}{\textbf{Inductive Link Prediction}} & 
        \multicolumn{3}{c}{\textbf{Node Classification }} \\
        &

        \multicolumn{2}{c}
        {\textbf{Wikipedia}} & 
        \multicolumn{2}{c}
        {\textbf{Reddit}} & 
        \multicolumn{2}{c}
        {\textbf{MOOC}} & 
        \multicolumn{2}{c|}
        {\textbf{LastFM}} &
        \multicolumn{2}{c}
        {\textbf{Wikipedia}} & 
        \multicolumn{2}{c}
        {\textbf{Reddit}} & 
        \multicolumn{2}{c}
        {\textbf{MOOC}} & 
        \multicolumn{2}{c|}
        {\textbf{LastFM}} &
        \textbf{Wikipedia} & \textbf{Reddit} &
        \textbf{MOOC} \\

        \midrule
        \textbf{Metrics} &
        \textbf{AP$\uparrow$} &
        \textbf{AUC$\uparrow$} &
        \textbf{AP$\uparrow$} &
        \textbf{AUC$\uparrow$} &
        \textbf{AP$\uparrow$} &
        \textbf{AUC$\uparrow$} &
        \textbf{AP$\uparrow$} &
        \textbf{AUC$\uparrow$} &
        \textbf{AP$\uparrow$} &
        \textbf{AUC$\uparrow$} &
        \textbf{AP$\uparrow$} &
        \textbf{AUC$\uparrow$} &
        \textbf{AP$\uparrow$} &
        \textbf{AUC$\uparrow$} &
        \textbf{AP$\uparrow$} &
        \textbf{AUC$\uparrow$} &
        \textbf{AUC$\uparrow$} &
        \textbf{AUC$\uparrow$} &  \textbf{AUC$\uparrow$} \\

        \midrule
        JODIE & 
        44.90 &
        40.01 &
        52.48 &
        49.38 &
        50.19 &
        50.06 &
        55.14 &
        57.91 &
        48.87 &
        44.86 &
        51.95 &
        48.84 &
        51.89 &
        50.51 &
        40.81 &
        35.76 &
        66.17 &
        55.17 &
        51.91 \\
        
        TGN & 
        54.94 &
        52.40 &
        58.64 &
        56.81 &
        52.82 &
        54.33 &
        55.38 &
        55.32 &
        54.10 &
        51.88 &
        55.70 &
        52.89 &
        52.73 &
        54.06 &
        57.87 &
        56.92 &
        36.11 &
        54.22 &
        50.24 \\
        
        TGAT & 
        46.99 &
        43.72 &
        66.71 &
        65.43 &
        52.06 &
        51.16 &
        51.05 &
        50.25 &
        47.74 &
        44.11 &
        62.90 &
        62.75 &
        51.88 &
        51.07 &
        50.91 &
        49.97 &
        54.03 &
        56.59 &
        45.53 \\
        
        GraphMixer &
        52.29 &
        49.26 &
        80.62 &
        79.01 &
        58.82 &
        57.45 &
        45.25 &
        42.26 &
        51.47 &
        48.35 &
        76.92 &
        75.39 &
        57.94 &
        56.75 &
        46.60 &
        44.43 &
        45.57 &
        54.32 &
        50.91 \\
        
        TCL & 
        \underline{81.00} &
        \underline{78.03} &
        54.24 &
        52.03 &
        70.10 &
        \underline{71.35} &
        50.84 &
        50.23 &
        80.73 &
        \underline{78.07} &
        54.31 &
        52.38 &
        69.40 &
        \underline{71.00} &
        51.10 &
        50.46 &
        \underline{69.72} &
        55.97 &
        54.25 \\

        DyGFormer & 
        80.60 &
        75.26 &
        84.55 &
        \underline{84.66} &
        \underline{72.93} &
        69.60 &
        85.33 &
        80.55 &
        \underline{81.38} &
        75.88 &
        85.27 &
        \underline{84.48} &
        \underline{71.99} &
        68.24 &
        83.53 &
        78.66 &
        61.68 &
        \underline{57.75} &
        55.70 \\

        \midrule

        TIGPrompt &
        52.97 &
        51.46 &
        56.19 &
        52.99 &
        52.15 &
        52.24 &
        53.37 &
        53.51 &
        52.64 &
        51.22 &
        54.58 &
        51.35 &
        51.92 &
        51.86 &
        56.04 &
        55.48 &
        46.49 &
        55.83 &
        56.16 \\

        DyGPrompt &
        55.22 &
        54.51 &
        54.63 &
        53.40 &
        51.26 &
        53.68 &
        54.97 &
        54.31 &
        53.71 &
        52.27 &
        53.43 &
        52.47 &
        53.31 &
        56.15 &
        52.69 &
        50.51 &
        59.24 &
        57.36 &
        48.27 \\

        \midrule
        \midrule
        
        TIGPrompt(D) &
        70.78 &
        65.22 &
        79.70 &
        79.94 &
        69.69 &
        69.15 &
        85.57 &
        80.70 &
        71.31 &
        65.72 &
        80.71 &
        80.13 &
        68.78 &
        63.84 &
        83.93 &
        78.95 &
        57.73 &
        57.62 &
        \underline{56.23} \\

        DyGPrompt(D) &
        76.07 &
        72.80 &
        \underline{85.47} &
        83.08 &
        52.33 &
        51.53 &
        \underline{86.19} &
        \underline{81.38} &
        75.81 &
        72.23 &
        \underline{85.54} &
        83.08 &
        52.24 &
        51.59 &
        \underline{84.57} &
        \underline{79.59} &
        66.53 &
        56.54 &
        47.37 \\

        \textbf{DDGPrompt} &
        \textbf{88.35} &
        \textbf{85.04} &
        \textbf{89.82} &
        \textbf{89.60} &
        \textbf{75.13} &
        \textbf{72.64} &
        \textbf{86.52} &
        \textbf{82.17} &
        \textbf{89.33} &
        \textbf{86.04} &
        \textbf{90.76} &
        \textbf{90.12} &
        \textbf{74.23} &
        \textbf{71.55} &
        \textbf{84.88} &
        \textbf{80.38} &
        \textbf{74.64} &
        \textbf{58.13} &
        \textbf{58.62} \\
        \bottomrule
    \end{tabular}
    }
\end{table*}

\(\bullet\) \textbf{Wikipedia} records the editing history and interaction behavior of users on the platform. The nodes represent different users or pages, and the edges represent the interaction information between users and pages.

\(\bullet\) \textbf{Reddit} is similar to Wikipedia, recording the interactions between users and different sub-posts on the Reddit website. The timestamp indicates the time of the interaction.

\(\bullet\) \textbf{MOOC} is derived from the learning activities and interactive behaviors on the online course platform, reflecting the students' participation in MOOC courses.

\(\bullet\) \textbf{LastFM} records the listening history of users on the music platform within a month, but there are no dynamic labels to indicate the user's status. It is a commonly used large dataset with 1.29 million interactions for link prediction on dynamic graphs.

\subsubsection{Task Setting}

We evaluate our method on two fundamental tasks in dynamic graphs: dynamic link prediction and dynamic node classification. For link prediction, we consider both transductive and inductive settings. In the transductive setting, the model has access to all nodes in the graph during training. In contrast, the inductive setting requires the model to predict links involving nodes that are entirely unseen during training. Due to the absence of dynamic labels, we only conduct experiments on the dynamic link prediction task using the LastFM dataset.

In our few-shot setting, we adopt a strict and challenging experimental protocol to evaluate the effectiveness of our method with extremely limited labeled samples. Specifically, we split the four datasets in the following ways: (1) First, we select the first 80\% of the interaction data in chronological order as the pretraining set \(G_{pre-train}\) for self-supervised pretraining. (2) Second, for the remaining 20\% of the data, we then select \textit{K} interactions as \textit{K}-shot training data to fine-tune in different downstream tasks. Similarly, we continuously select \textit{K} samples as the validation set for fine-tuning, denoted as \(G_{fine-tune}\) and \(G_{val}\). Finally, for all the remaining data, we use it as the test set \(G_{test}\) for downstream tasks. We just select \textit{K} = 70 as the 70-shot scenario for all downstream tasks in the main experiment, which accounts for only 0.05 percent of all interactions for the smallest dataset, Wikipedia.

Besides, for the dynamic node classification task, we first ensure that at least one node is selected in each class when constructing the fine-tuning set and the validation set.

\subsubsection{Baselines} We compare DDGPrompt with the following two types of baseline models. (1) \textbf{Dynamic graph learning}: JOIDE \cite{jodie}, TGAT \cite{TGAT}, TGN \cite{tgn}, GraphMixer \cite{graphmixer}, TCL \cite{tcl}, DyGFormer \cite{dygformer}. (2) \textbf{Dynamic graph prompt}: TIGPrompt \cite{tigprompt}, DyGPrompt \cite{DyGPrompt}.

For TIGPrompt and DyGPrompt, we adopt the best implementation setting reported in their original papers by default, i.e., using TGN as the backbone. Since the code of TIGPrompt and DyGPrompt are not available, we implemented them based on the descriptions provided in their respective papers. Additionally, to ensure a fair comparison, we also integrated the prompts into several recent dynamic graph methods and reported the corresponding results in the main experiments.

\subsubsection{Evaluation Metrics}Following the previous work, we use Average Precision (AP) and Area Under the Receiver Operating Characteristic Curve (AUC-ROC) as evaluation metrics  for the dynamic link prediction. For the dynamic node classification, we only use AUC-ROC due to label imbalance.

\subsubsection{Hyperparameter Settings and Implementation}
We build and evaluate all baselines using the DyGLib \cite{dygformer} benchmark. In pretraining, we set all baseline epochs to 100, the learning rate to 0.0001, do not deploy early stopping, and save the best model. The temperature parameter \(\tau\) of contrastive learning loss in pretraining is set to 0.2. For all downstream tasks, we load and freeze the pre-trained models and use its baseline default parameters in DyGLib (usually optimal) with 20 early stopping to tune the task classifier. We set the learning rate for downstream tasks to 0.0001 for dynamic link prediction and 0.001 for dynamic node classification. We run 5 times with different seeds and calculate the average of all results as the final performance of the model. All experiments were conducted on a Linux server with a single GPU (GeForce RTX 3090). 

\subsection{Analysis of Main Experiments Performance}
\label{5.2}
We evaluate all methods in four datasets using dynamic link prediction, and in the dynamic node classification task on Wikipedia, Reddit, and MOOC. In addition, for fairness we implement TIGPrompt and DyGPrompt on TCL and DyGFormer respectively, and compare their performance with our DDGprompt. We use the first letter to represent the selected backbone model for prompt. For example, TIGPrompt (D) represents TIGPrompt based on the DyGFormer. We use DyGFormer as the default backbone of our method. The experimental results are shown in Table \ref{tab:main_shiyan}. 

We bold the best result and underline the second-best result on each dataset. From the results, we have the following observations.

(1) DDGPrompt achieves the best performance in most datasets for different downstream tasks. Specifically, on dynamic link prediction, DDGPrompt outperforms all baselines in the Wikipedia, Reddit, and LastFM datasets. For the Wikipedia, DDGPrompt improves the backbone by 11.37\% and 13.29\% on AP and AUC, respectively. On Reddit and MOOC, DDGPrompt still improves the selected backbone. The limited improvement on Reddit is attributed to the overall performance bottleneck. In the dynamic node classification task, DDGPrompt also significantly improves the performance of the selected backbone. The above results prove that DDGPrompt can improve the performance of pre-trained models on different downstream tasks, demonstrating the effectiveness of our proposed method.

(2) DDGprompt outperforms TIGPrompt and DyGPrompt when using the same backbones. We take the AP evaluation indicator as an example. When the backbone is DyGFormer, our DDGPrompt outperforms TIGPrompt and DyGPrompt by 18.37\% and 8.23\% in link prediction for LastFM. For the node classification, DDGPrompt outperforms TIGPrompt and DyGPrompt by 8.36\%/6.48\%/2.4\% and 6.51\%/2.95\%/14.48\% respectively. The above observations demonstrate that our proposed DDGprompt outperforms existing simple dynamic graph prompting methods \cite{tigprompt,DyGPrompt} by more comprehensively tuning the temporal embeddings of nodes according to downstream tasks.

(3) In addition, different backbones also has a great impact on the final performance. For the dynamic link prediction , we notice that when DDGPrompt selects the best backbone DyGFormer, although it has a certain improvement on MOOC, the final result is inferior to the result when TCL is the backbone. This may be because the performance of the DyGFormer is seriously weaker than TCL in the current experimental setting.

\subsection{Analysis of Model Performance on Sparse Interaction Data}

In this section, we focus on scenarios with limited interaction data. By filtering and analyzing datasets with sparse interaction nodes, we aim to demonstrate how DDGPrompt can effectively adapt and perform well under the more stringent condition. Specifically, we filter all nodes that have fewer than a specified interaction threshold, retaining only the corresponding interactions for these nodes from the previous datasets, thereby limiting both the sample number and the interaction frequency. For LastFM, we set the interaction threshold to 1000, while for the other three datasets, the threshold is set to 100. This results in the labeled data for each dataset being at most half of the 70-shot scenario in Section \ref{5.2}. We show the performance of each baseline under sparse interaction data in Table \ref{tab:few_interact_exp}.

We can observe that under this setting, DDGPrompt achieves the best performance on all datasets, highlighting the robustness of our method. It can maintain high performance even when only a small number of interactions are available. This analysis not only emphasizes the advantages of our model but also significantly contributes to enhancing its applicability in real-world scenarios involving cold starts.

\begin{figure}[htbp]
    \centering
\includegraphics[width=0.47\textwidth]{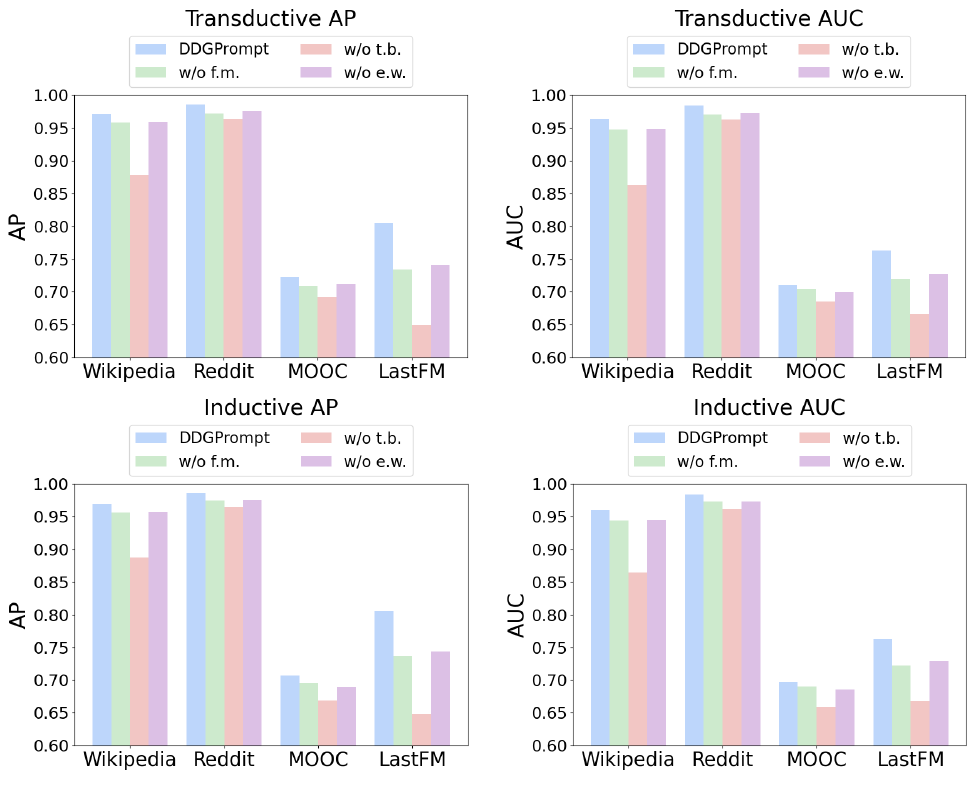}
    \caption{Ablation Studies of prompt components}
    \label{fig:Ablation}

\end{figure}
\subsection{Analysis of Model Components}

In this section, we perform ablation experiments on DDGPrompt to verify the effectiveness of each prompt component. We explore three variants based on DDGPrompt and compare them with DDGPrompt. Fig. \ref{fig:Ablation} shows the corresponding results of these variants on the dynamic link prediction task in four datasets. The variable \textbf{w/o t.b.} refers to DDGPrompt with the temporal bias prompt removed. Variant \textbf{w/o e.w.} and variant \textbf{w/o f.m.} represent without edge weight prompt and feature mask prompt, respectively.

We can see that all three proposed prompts are advantageous for dynamic link prediction, and they demonstrate a similar pattern across different datasets. For most datasets, the model performance degrades the most when the temporal bias prompt is absent. Additionally, the effects of the prompts vary across different datasets. In particular, for the Wikipedia and LastFM datasets, each prompt contributes significantly to overall performance improvement. Overall, these observations suggest that the components of our proposed DDGPrompt can adjust node temporal embeddings from different perspectives to better fit specific downstream tasks.

    

\subsection{Analysis of Model Hyperparameters}

\begin{figure}[htbp]
    \centering
    
    \includegraphics[width=0.45\textwidth]{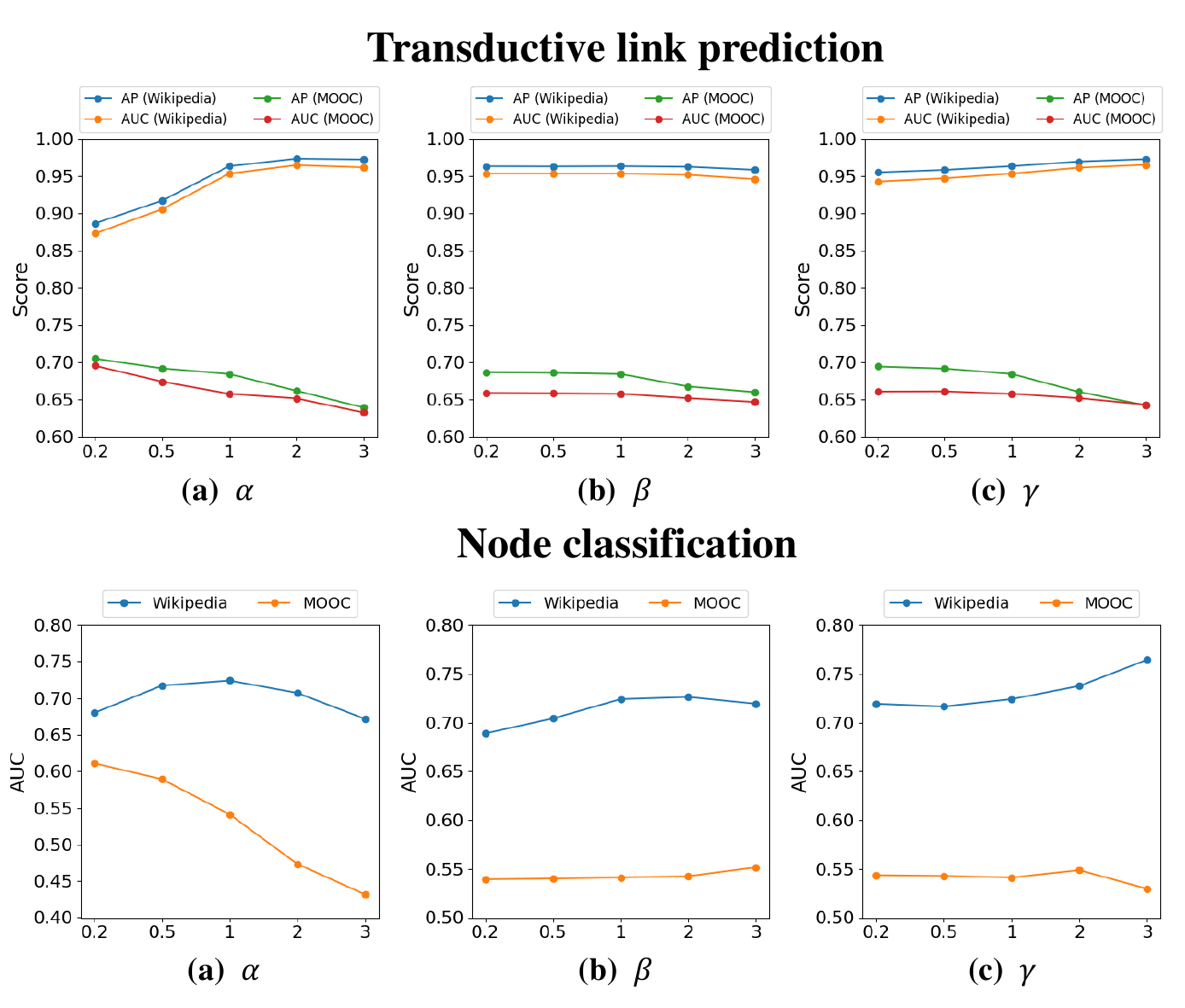}
    \caption{Performance under different hyperparameters on transductive link prediction}
    \label{fig:para}

\end{figure}

Then, we analyze the hyperparameter of DDGPrompt. Specifically, we evaluate the weights of each prompt when added to the original node expression feature matrix under the transductive link prediction, i.e. \(\alpha\), \(\beta\), \(\gamma\) in Eq. \ref{eq:add_all_prompt}. We tested the results of the three hyperparameter values on Wikipedia and MOOC, reflecting the impact of different prompts on the backbone across various datasets. The results for different downstream tasks are presented in Fig. \ref{fig:para}.

We can observe that in the transductive link prediction, as we increase the weight \(\alpha\) of the temporal bias prompt, the performance of DDGPrompt continuously improves in the Wikipedia dataset while declining in the MOOC. This indicates a larger time gap in the Wikipedia dataset, while the opposite holds for the MOOC. The weight \(\gamma\) of the feature mask prompt shows a similar trend.


\subsection{Analysis of Model Scalability}

In this section, we analyze the scalability and effectiveness of our proposed DDGPrompt on different backbones. We can notice the performance and improvement of DDGPrompt on different state-of-the-art backbone networks in Table \ref{tab:main_shiyan}. In the previous section, we can notice that although existing prompt works improve the performance of some tasks, their performance is often disappointing when faced with different datasets. In contrast, we observe in Table \ref{tab:main_shiyan} that although different backbone networks have an impact on the final results, our DDGPrompt improves the backbone that do not use prompt. In particular, for the current best backbone DyGFormer, our DDGPrompt has the greatest improvement. This proves the scalability and effectiveness of our proposed DDGPrompt on different backbones.

\subsection{Analysis of Node Feature Distribution}

\begin{figure}[htbp]
    \centering
    
    \includegraphics[width=0.46\textwidth]{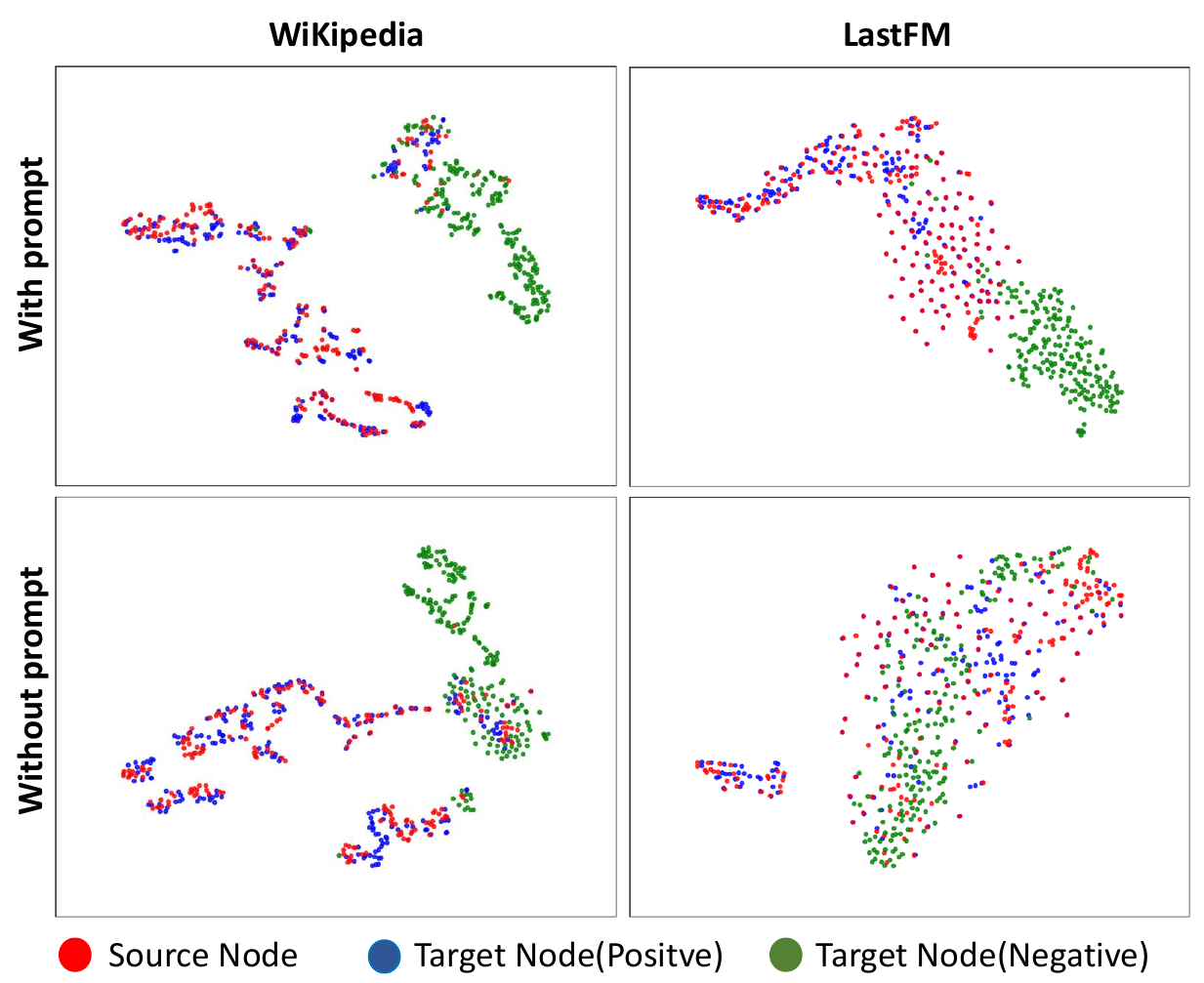}
    \caption{Visualization of node feature distributions with and without DDGPrompt on dynamic link prediction (DyGFormer as the backbone)}
    \label{fig:node_feature_distribution}

\end{figure}

To better demonstrate the effectiveness of our method, we visualize the node feature distribution with and without DDGPrompt on DyGFormer. Specifically, we randomly sample 200 node pairs from the test sets of Wikipedia and LastFM, and perform dimensionality reduction for visualization. Fig.~\ref{fig:node_feature_distribution} presents the link prediction results, where each point represents a node involved in either a positive or negative sample pair. We expect the node representations of positive pairs to be close to each other, while those of negative pairs should be far apart in dynamic link prediction task. As shown in Fig.~\ref{fig:node_feature_distribution}, after applying DDGPrompt, positive and negative pairs become more clearly separable in the embedding space, and the nodes within positive pairs are more tightly clustered. This indicates that the model, guided by DDGPrompt, is better at distinguishing whether two entities are likely to interact, demonstrating its ability to enhance semantic discrimination in downstream tasks.

\subsection{Analysis of Time and Memory Consumption}

In addition, we test the training overhead of existing dynamic graph prompt methods, including time and memory consumption in the main experimental setting. We test on Wikipedia and the large dataset LastFM and calculate the average results of five experiments. Table~\ref{lp_overhead} and Table~\ref{nc_overhead} show the experimental data of dynamic link prediction and dynamic node classification tasks respectively.
\begin{table}[h]
    \centering
    \caption{Dynamic graph prompt methods training overhead for dynamic link prediction}
    \begin{tabular}{lcccc}
        \toprule
         &
        \multicolumn{2}{c}{\textbf{Time}(s)} &
        \multicolumn{2}{c}{\textbf{Memory}(MB)}
        \\ & Wikipedia & LastFM & Wikipedia & LastFM \\ \midrule
        TIGPrompt & 98 & 954 & 1010 & 2582 \\
        DyGPrompt & 126 & 1096 & 946 & 2542 \\
        DDGPrompt & 102 & 1098 & 894 & 2480 \\
        \bottomrule
    \end{tabular}

    \label{lp_overhead}
\end{table}

\begin{table}[h]
    \centering
    \caption{Dynamic graph prompt methods training overhead for dynamic node classification}
    \begin{tabular}{lcccc}
        \toprule
         &
        \multicolumn{1}{c}{\textbf{Time}(s)} &
        \multicolumn{1}{c}{\textbf{Memory}(MB)}
        \\ & Wikipedia & Wikipedia \\
        \midrule
        TIGPrompt & 24 & 992 \\
        DyGPrompt & 28 & 932 \\
        DDGPrompt & 18 & 888 \\
        \bottomrule
    \end{tabular}
    
    \label{nc_overhead}
\end{table}

We can observe that the training costs of the three dynamic image prompting methods are close to each other. Although our proposed DDGPrompt involves a weighted combination of multiple prompt matrices, it still has a small training time overhead because the time complexity is still linear. In terms of memory overhead, TIGPrompt has the largest memory consumption due to its inclusion of transformers, followed by DyGPrompt with a dual MLP network. DDGPrompt has the smallest memory consumption due to the least trainable parameters.

\section{Conclusion}

In this paper, we propose DDGPrompt, a novel data-centric prompting framework for dynamic graphs. It bridges the gap between pre-trained models and downstream tasks by adaptively refining node embeddings based on interaction patterns through three prompts. Extensive experiments on four dynamic graph datasets under few-shot settings show that DDGPrompt outperforms both traditional baselines and existing prompt-based methods.

\begin{acks}
This work was supported by the National Key Research and Devel
opment Program of China (No.2023YFC3303800)
\end{acks}

\bibliographystyle{ACM-Reference-Format}
\balance

\bibliography{ref}

\end{document}